# Towards life cycle identification of malaria parasites using machine learning and Riemannian geometry

Arash Mehrjou[*]
[*]*Max Planck Institute for Intelligent Systems, Tübingen, Germany*
e-mail: arash.mehrjou@tuebingen.mpg.de

**Abstract**

Malaria is a serious infectious disease that is responsible for over half million deaths yearly worldwide. The major cause of these mortalities is late or inaccurate diagnosis. Manual microscopy is currently considered as the dominant diagnostic method for malaria. However, it is time consuming and prone to human errors. The aim of this paper is to automate the diagnosis process and minimize the human intervention. We have developed the hardware and software for a cost-efficient malaria diagnostic system. This paper describes the manufactured hardware and also proposes novel software to handle parasite detection and life-stage identification. A motorized microscope is developed to take images from Giemsa-stained blood smears. A patch-based unsupervised statistical clustering algorithm is proposed which offers a novel method for classification of different regions within blood images. The proposed method provides better robustness against different imaging settings. The core of the proposed algorithm is a model called Mixture of Independent Component Analysis. A manifold based optimization method is proposed that facilitates the application of the model for high dimensional data usually acquired in medical microscopy. The method was tested on 600 blood slides with various imaging conditions. The speed of the method is higher than current supervised systems while its accuracy is comparable to or better than them.

# 1. Introduction

Automatic diagnosis of malaria is a paradigm focusing on translating the medical expertise of malaria diagnosis into computer platform. Malaria is a serious disease caused by genus *Plasmodium* that invades Red Blood Cells (RBCs). There exist four species of malaria parasite and four life-stages. Morphologically differentiable life-stages are named *ring*, *trophozoite*, *schizont* and *gametocyte*. Different life-stages for different malaria species are represented in Figure 1a, Figure 1b, Figure 1c, and Figure 1d corresponding to *P. Flaciparum*, *P. Vivax*, *P. Ovale*, and *P. Malariae* respectively. Fast and accurate diagnosis is essential for curing patients. Unfortunately, Malaria is mostly common in low-income developing African countries where even preliminary medical cares are not publicly available. The official organizations have reported about 200 million new patients and over half million deaths yearly. In this paper, the aim is to firstly detect the presence of the parasite and secondly determine its life-stage.

There are two major methods for detection of malaria parasites: *microscopy* and *rapid diagnostic tests* (RDTs). Microscopy is preferable over RDTs with regard to costs and is recognized as the gold standard in parasite quantification [1]. More information like species, life-stages of parasites and drug influence monitoring are obtainable from microscopy method if needed. On the other hand, microscopy requires skilled operators, maintenance of equipment, and capital investment. Microscopy is time consuming especially in regions with high caseload and this issue usually results in variable reports [2] and poor performance [1, 3]. Some reports show that the accuracy of microscopy results obtained by human operators is surprisingly low and only 51% of parasite quantification tasks are acceptable [4]. The cure is to minimize the human role in the diagnosis process. Imaging CCD sensors have made digital microscopy possible by producing high resolution images form the slides under microscope lens [5]. Different image features like color, texture and shape have been used in several researches on vision-based detection and quantification of malaria parasites [6, 7]. Current vision-based malaria diagnosis methods are mostly supervised and must deal with issues like staining quality, illumination variability, image pre-processing, feature extraction, feature selection , and classification schemes [6]. Considering the high infection rates and huge amount of image data, we need a fast accurate method with minimum human intervention that is robust against different sources of variations. Variations happen in several stages of the diagnosis process and threaten the performance of the overall system. The Sources of variations can be identified in each diagnosis stage as follows:

- **Slide preparation**: Blood samples are taken from patients and stained with Giemsa solution. The color of each blood constituent element changes by a small variation in the PH of the Giemsa solution [6].

- **Image Acquisition**: This task requires some interfaces between the camera, the microscope and the computer. The complete diagnosis system should be affordable for low-income countries around the world. Therefore, we need to build an economical system for this task instead of using expensive pre-built systems. Hence, we bought low-priced devices for this project which are shown in Figure 2a, Figure 2b, and Figure 2c. An electronic circuit is developed to drive the motors and make the microscope controllable via a serial port of a laptop computer. This system automates the image acquisition of the blood slides. The final system is depicted in Figure 2d. The whole accessories of this system are listed in Table 1. It is clear that each part of this system may be bought from different providers with different characteristics. For example, dissimilar microscopes may cause different illumination conditions in blood slide images as shown in Figure 3. The same issue is likely for different CCD cameras. These will be the source of variations in illumination, color map, resolution, scale and etc.

- **Segmentation**: In some methods, it is necessary to first segment Red Blood Cells (RBC) before further diagnostic processing. Overlapping and misshaped cells may affect the accuracy of the segmentation algorithm and becomes a source of variation.

- **Parasite detection**: The Giemsa staining process makes several blood components observable. The stained pixel groups are categorized into parasites (4 species and 4 life-stages) and non-parasites (White Blood Cells (WBCs), platelets, and artefacts). A detected stained pixel group should be further processed to determine if it is a parasite.

- **Identification of parasite species and life-stages**: Overlapping of parasite species in some life-stages makes the classification task very difficult and often impossible even by human experts. Therefore, we think it is reasonable to focus on life-stage classification which has important applications in prescription by determining how much the infection is progressed. In this paper, we focus on confronting the variations and enhancing the parasite detection algorithms in an unsupervised manner. The algorithm detects parasites and processes them to identify their maturity. This will be achieved by exploiting the information within rectangular patches of blood images and taking advantage of color, texture and scale in a single unified algorithm. The rest of the paper is organized as follows: Section 2.1 defines the mathematical model and section 2.2 proposes an optimization method to solve this model over data. Sections 3.1 and 3.2 present the use of this mathematical framework for invariant diagnosis in pixel level and parasite level respectively. Finally, the paper is concluded in section 4 and future possible research directions are suggested.

## 2. Mathematical formulation

In this section, we first brief the notation and formulation of the proposed unsupervised method. Second, we discuss the manifold optimization solution for this problem and manifest its advantages.

### 2.1 Mixture of ICAs model

Sparse codes are known to be a suitable representation for texture information [8]. It is well known that ICA extracts sparse features from natural images [9]. Decomposing data by a single ICA assumes that the whole data distribution is adequately described by one coordinate frame. However, this is obviously incorrect for many tasks, e.g. when images consist of different regions each with different statistical characteristics. Therefore, assuming a single ICA model for images fails to capture most of the data statistics. The remedy that we propose here is to use a mixture model partially inspired by [10] that enhances the modeling capacity of a single ICA. We propose a mathematical solution which facilitates the use of this relatively complex model. A single ICA as $k^{th}$ component of a MoICA model is described by

$$\mathbf{x}_k = \mathbf{A}_k \mathbf{s}_k + \mathbf{n}_k, \tag{1}$$

where subscript $k$ represents the association to $k^{th}$ mixture component. Vector $\mathbf{x}_k$ is the $M-dimensional$ mixture vector, also called sensor vector. Vector $\mathbf{s}_k$ is the $L_k-dimensional$ source vector and $\mathbf{n}_k$ is the Gaussian noise. Moreover, $\mathbf{A}_k$ is the $M \times L_k$ mixing matrix for describing the mixing process of $k^{th}$ coordinate frame. Sensor and source dimensions are assumed to be equal for all mixture components that implies $L_k = L$ and $M_k = M$. For simplicity, we consider square noiseless case in which $M = L$ and $cov(\mathbf{n}_k) = 0$. Therefore, we have

$$p(\mathbf{x}_k|\mathbf{A}_k,\boldsymbol{\theta}_k) = \frac{1}{|det(\mathbf{A}_k)|}p(\mathbf{s}_k|\boldsymbol{\theta}_k), \tag{2}$$

where vector $\boldsymbol{\theta}_k$ aggregates the set of parameters for all sources of $k^{th}$ component. Due to the independence assumption, the source vector probability density of each mixture component is written as

$$p(\mathbf{s}_k|\boldsymbol{\theta}_k) = \prod_{i=1}^{L_k} p(s_{k,i}|\boldsymbol{\theta}_{k,i}), \tag{3}$$

where $k,i$ represents the $i^{th}$ source of $k^{th}$ component and $\boldsymbol{\theta}_{k,i}$ represents the parameters of this source. In this paper, we have extended the fixed source models used in [11, 12] by considering each source as a MoG which allows a wide variety of distributions to be modeled. Thus, (3) becomes

$$\begin{aligned}
p(\mathbf{s}_k|\boldsymbol{\theta}_k) &= \prod_{i=1}^{L_k}\sum_{q_i=1}^{m_i} p(q_i|\boldsymbol{\pi}_i,k)\, p(s_{k,i}|\boldsymbol{\theta}_{k,i}) \\
&= \prod_{i=1}^{L_k}\sum_{q_i=1}^{m_i} \pi_{i,q_i}\, \mathcal{N}(s_{k,i};\mu_{i,q_i},\sigma_{i,q_i}),
\end{aligned} \tag{4}$$

where subscript $k$ has been dropped from the parameters belonging to $k^{th}$ component, because it is clear from the context. Therefore, $q_i$ is the index of the Gaussian component contributing to $i^{th}$ source of $k^{th}$ ICA model. Similarly, $m_i$ is the number of Gaussian components used to model $i^{th}$ source. The likelihood of the I.I.D. dataset $\mathbf{X} = \{\mathbf{x}^1, \mathbf{x}^2, \dots, \mathbf{x}^T\}$ given the model parameters $\boldsymbol{\varphi}_k = \{\mathbf{A}_k, \boldsymbol{\theta}_k\}$ can now be written by integrating over latent variables $\{\mathbf{s}_k, \mathbf{q}_k\}$ as

$$p(\mathbf{X}|\boldsymbol{\varphi}_k,k) = \prod_{t=1}^{T}\sum_{\mathbf{q}_k=1}^{\mathbf{m}_k} \int p(\mathbf{x}^t, \mathbf{s}_k^t, \mathbf{q}_k^t|\boldsymbol{\varphi}_k,k)\, d\mathbf{s}_k, \tag{5}$$

where $d\mathbf{s}_k = \prod_i ds_{k,i}$, $\mathbf{q}_k = [q_1, q_2, \ldots, q_{L_k}]$, and $\mathbf{m}_k = [m_1, m_2, \ldots, m_{L_k}]$. Finally, the probability of generating dataset **X** from a $K - component$ mixture model is given by:

$$p(\mathbf{X}|\mathbf{\Phi}) = \sum_{k=1}^{K} p(k|\varphi_0) p(\mathbf{X}|\boldsymbol{\varphi}_k, k), \tag{6}$$

where $= \{\varphi_0, \boldsymbol{\varphi}_1, \ldots, \boldsymbol{\varphi}_K\}$. Parameter $\varphi_0$ represents the probabilistic membership of the mixture model components. Substituting (2), (4), and (5) in (6) yields a maximum likelihood model. This can then be learned through an iterative process such as Expectation Maximization algorithm [13] or gradient descent [11, 12]. In this paper, however, a manifold optimization technique is used which is described in next section.

## 2.2  Manifold optimization process

Optimizing the problem specified by (6) is a difficult task in general. In traditional ICA algorithms, constraints are imposed between subsequent iterations or penalize the contrast function in case of violation [14]. Nevertheless, these attempts do not guarantee the perpetual satisfaction of the constraints. Therefore, the Riemannian optimization is adopted here instead of optimizing over the Euclidean space. This method forces the solutions to be *sticked* to the desirable subspace of the entire space. Manifold method will facilitate the relaxation of the orthogonality constraint which is present in most of the current ICA algorithms and ensures good convergence characteristics [14, 15]. This is expected to improve the obtained solutions due to the increased degrees of freedom. The optimization method used in this paper, i.e., limited Broyden-Fletcher-Goldfarb-Shanno (L-BFGS), is then intended for oblique manifold ($\mathcal{OB}$) which is a manifold of matrices with normalized columns.

# 3. Automatic diagnosis

In this section we shall show the application of the proposed mathematical formulation described in section 2 for the diagnosis procedure. We first describe the advantages of using this method in stained/non-stained pixel classification task and then show its application in identification of parasite life-stages.

## 3.1 Stained/non-stained pixel classification

In the first stage of the diagnosis process, every pixel should be tagged as stained or non-stained. Stained pixels may belong to parasites, WBCs, platelets, or artefacts. The best result of the state of the art methods have been achieved by [16] where a minimum error Bayesian decision making is performed over RGB images. The labeled data has been achieved by a manual double-thresholding morphological operator over about 100 images to provide sufficient training data. The process clearly takes huge amount of time and should be done for each Giemsa-staining and CCD imaging settings. Although we do not use a supervised method to detect stained pixels in this paper, we have developed a software package in MATLAB as shown in Figure 4a and Figure 4b for double-thresholding task to obtain the ground-truth data. We have developed another software package as shown in Figure 4c and Figure 4d that allows the physicians to annotate information of parasite species and life-stages on blood images for evaluation of diagnostic softwares or educational aims. The main drawbacks of the supervised approaches as one used by [6] are :

1. **Time**: Supervised methods will take experts a lot of time to determine either each pixel is stained or not for about 100 images by hand. This task is also prone to human errors.

2. **Considering single pixels**: The probability of finding a stained pixel near a colony of previously marked pixels is higher than near random pixels. This valuable information is not used in the supervised approach [6].

3. **Ignoring color in marking process**: In the manual double-thresholding process, the obtained color images from malaria blood slides are converted to grayscale images and 3-channel information is missed. This issue will definitely harm the performance of the entire diagnosis system because of its fundamental role.

4. **Susceptible to uneven illumination:** Illumination change over a single slide or multiple slides will affect the classification process [6]. This will require a pre-processing method for illumination normalization before further processing. It will take extra amount of time and its results may be inconsistent for different microscopes and imaging devices.

Our proposed method deals with these issues in a unified unsupervised framework. We consider a generative model like (6) for the formation of image patches. A diverse collection of 100 images with different slide preparation and imaging conditions are put in a folder from which $1000000$ random $32\times32$ windows are selected to form a collection of $32\times32\times3$ color patches. Each patch is then vectorized into a $3072\times1$ matrix by cascading its 3 color channels. This process produces a dataset **X** of size $3072\times1000000$. Data dimension is then reduced to $256$ by PCA whitening. Whitening will speed up the optimization process by considerable reduction in the number of free parameters [9]. A $2-$ component MoICA model is set up such that each ICA component has a factorial density with 256 sources. The model (6) is then trained with dataset **X**. The optimization is performed by L-BFGS algorithm on the oblique manifold ($\mathcal{OB}$). Using manifold optimization method gives us the opportunity for taking ad-vantage of *Minibatch* trick. This trick will allow us to optimize fractions of the dataset in each run instead of the whole dataset and will speed up the entire process considerably. Some of the features, which are the columns of mixing matrix $\mathbf{A}_k$ reshaping into rectangular patches, are learned from the aforementioned collected dataset $\mathbf{X}$ and shown in Figure 5. The features learned by the MoICA component corresponding mainly to the background cells are shown in Figure 5a. Learned features of the other component corresponding to the foreground pixels are represented in Figure 5b. Here, foreground pixels are stained objects in thin blood smears. The variation of color, position and angle in the learned features provides desirable robustness against these variations in analysis of malaria slides. Another dataset $\mathbf{X}'$ consisting of $1000000$ patches randomly picked from 100 images other than those used for training is constructed. Then, the Bayes posterior probability

$$p(k|\mathbf{x}^t) = \frac{p(\mathbf{x}^t|k)p(k)}{\sum_{k=1}^{K} p(\mathbf{x}^t|k)p(k)} \qquad (7)$$

determines how likely data point $\mathbf{x}^t$ belongs to the ICA component 1 or 2 of the MoICA model. For example, for the blood slide image shown in Figure 6a, superimposed red patches in Figure 6b are the data points that maximize the above expression for the ICA component corresponding to the foreground stained pixels. The patches related to the other ICA component that captures the statistics of the uninfected parts of the image are not shown in Figure 6b for visual clarity. As clearly seen, the red patches are mostly concentrated around stained pixel groups which are the regions of interest in this stage of the diagnosis process.

To compare the result of our unsupervised stained pixel detection method with the supervised method of [6], 200 images from our blood smear dataset were selected. We used 100 images for training the supervised algorithm. Then, both algorithms were tested over the rest 100 images. The results are represented in Table 2. As was expected, our method performs as well as the supervised method but marginally below it. This is due to the fact that patches impose rectangular borders on the stained pixel groups. Notice that the final aim of a malaria diagnosis system is to detect parasites not the stained pixels. Thus, the result of the unsupervised method in this stage is completely justified by its advantages over supervised methods in time consumption and the invariance gained by learned features. The stages like *uneven illumination correction*, *color normalization*, and *supervised marking of the stained pixels*, are all removed in our method which reduces the required time of the detection process significantly. This advantage is due to the fact that our method learns meaningful features of the stained pixel groups instead of using hand crafted pre-designed features.

### 3.2 Parasite life-stage determination

In this section, maturity of parasites is of interest. This information will be helpful to monitor the progress of the infection. The algorithm is described for ring life-stage and is treated the same for other three life-stages. As shown in Figure 6b the patches which are associated to the stained pixels, roughly delineate the border of the stained pixel groups. We call each stained pixel group determined by section 3.1 an *island* or *stained object*. Each island can be one of the four life-stages or one of the members of the set {WBC, platelet, artefacts}. WBCs are initially removed due to their larger size compared to the other blood elements. Therefore, the task is now a $6-$class classification. Here is the point that we let human intervention to play its role. We solve a $6-$component MoICA model of section 2.1 for the patches belonging to the islands using mathematical method described in section 2.2. Then, we show the learned features to the operator (who is not necessarily a highly trained expert microscopist) and ask him/her to mark those features which are visually correlated to a specific type of stained object. For example, the representative features for rings are marked by a non-expert operator in Figure 7. Assume that we let the MoICA model learn $L$ features for each ICA component. This is another advantage of the proposed model and manifold optimization solution compared to the classic square ICA problems where number of features must be equal to the dimension of input data. Then, $L$ learned features are partitioned into $S$ subspaces such that each subspace describes the features which are visually correlated to a specific type of stained objects. Note that, this process does not require the features of each subspace to be mutually exclusive. A feature can be used to represent two different stained objects. After determining the $S$ subspaces of features, we need to associate each island, as a set of several patches, to one of these subspaces. This will be done by projecting each patch of the island onto the all subspaces and determine which subspace describes that patch best. To formalize this process, imagine that $\mathbf{x}_{i,p}$ is the $i^{th}$ patch from $p^{th}$ island. According to (1), in a noiseless case, each patch is described by a linear combination of features

$$\mathbf{x}_{i,p} = \sum_{1 \leq l \leq L} \mathbf{A}_l s_{i,p,l} \qquad (8)$$

where $i, p$ represents the $i^{th}$ patch from $p^{th}$ island. The vector $\mathbf{A}_l$ is the $l^{th}$ column of mixing matrix $\mathbf{A}$ which is named "feature" in this context. It is emphasized that (8) is written for the component of the MoICA model which corresponds to the stained objects {parasites, WBCs and artefacts}. Note that $s_{i,p,l}$ is the result of projecting the patch $\mathbf{x}_{i,p}$ over the direction of $l^{th}$ feature. To calculate the projection of $\mathbf{x}_{i,p}$ over $s^{th}$ subspace, we integrate the projections over the features of that subspace in a single value as

$$e_{i,p,s} = \sqrt{\sum_{l \in s^{th}} s_{i,p,l}^2} \qquad (9)$$

where $e_{i,p,s}$ represents the result of projecting the $i^{th}$ patch of the $p^{th}$ island onto the $s^{th}$ subspace. The class of patch $\mathbf{x}_{i,p}$ is determined as the maximizing $s$ for $e_{i,p,s}$ in (9). Finally, each island is identified as class $s$ if the majority of its patches are related to subspace $s$. For example, after determining the subspace of ring-shaped parasites, the majority of the patches around ring-shaped islands are found associated to that subspace and the majority voting method assigns that island to the class of ring-shaped parasites. A flowchart that summarizes the whole process is shown in Figure 8.

As discussed, the proposed patch-based approach uses local learned features. Thus, to improve the accuracy of the parasite life-stage classification, a vector of shape-based features {area, perimeter, compactness} has been used to augment the patch-based features. The comparison between the performance of the proposed unsupervised method used in this paper and the supervised approach used by Tek in [6] is presented as two confusion tables in Table 3a and Table 3b. This is achieved for 600 images obtained from different fields of our 20 available Giemsa-stained blood smears. The initials represent the following items: R (Ring), T (Trophozoites), S (Schizonts), G (Gametocytes), P (Platelets), A (Artefacts). As marked in Table 3a, our method shows significant improvement in discrimination between ring and trophozoite life-stages. Also, the recognition of artefacts and platelets are more accurate in our method compared to the completely supervised case. Other entries reveal that our method produces comparable results to the totally supervised algorithm.

## 4. Conclusion

Contributions of this paper are both practical and theoretical. We have manufactured a controllable motorized microscope and designed its serial interface to computer with an electronic circuit. On the software side, we have developed peripheral applications to obtain the ground-truth dataset in both pixel level and object level. This large labeled dataset can also be used for training malaria expert microscopists or in further research on malaria automatic diagnosis. As theoretical contributions, we have proposed a MoICA model for classification of elements in thin stained blood smears. We have used a novel manifold optimization solution to MoICA problem with flexible MoG sources for its ICA components. This optimization method makes the solution of such a complicated flexible model achievable. We have clearly specified the level in which human operator can intervene in parasite life-stage classification. The classification result is comparable and in some cases better than completely supervised case. Our unsupervised method is more time-efficient and more robust to variations in the diagnosis procedure. In this paper, parasite life-stages are processed and discriminated. Future studies may try to develop algorithms to combine this unsupervised idea with specialized hand crafted features to obtain satisfying results in classification of malaria parasite species as well as life-stages. If large number of blood smears are available, statistical modeling may be applicable to obtain a model of how different species and life-stages of malaria parasites found together in a blood sample. This model would work as a general prior rule to augment future diagnosis systems and also help achieve better understating of how malaria infection behaves within human body.

# References

[1] Wongsrichanalai C, Barcus MJ, Muth S, Sutamihardja A, Wernsdorfer WH. A review of malaria diagnostic tools: microscopy and rapid diagnostic test (RDT). Am J Trop Med Hyg 2007; 77(6 Suppl): 119-127.

[2] O'Meara WP, Barcus M, Wongsrichanalai C, Muth S, Maguire JD, Jordan RG, Prescott WR, McKenzie FE. Reader technique as a source of variability in determining malaria parasite density by microscopy. Malar J 2006; 5(1): 118.

[3] Zhu H, Sencan I, Wong J, Dimitrov S, Tseng D, Nagashima K, Ozcan A. Cost-effective and rapid blood analysis on a cell-phone. Lab Chip 2013; 13(7): 1282-1288.

[4] Frean J, Perovic O, Fensham V, McCarthy K, Gottberg Av, Gouveia Ld, Poonsamy B, Dini L, Rossouw J, Keddy K. External quality assessment of national public health laboratories in Africa, 2002-2009. Bull WHO 2012; 90(3): 191-199.

[5] Linder E, Lundin M, Thors C, Lebbad M, Winiecka-Krusnell J, Helin H, Leiva B, Isola J, Lundin J. Web-based virtual microscopy for parasitology: a novel tool for education and quality assurance. PLoS Negl Trop Dis 2008; 2(10): e315.

[6] Tek FB, Dempster AG, Kale I. Computer vision for microscopy diagnosis of malaria. Malar J 2009; 8(1): 153.

[7] Mehrjou A, Abbasian T, Izadi M, editors. Automatic Malaria Diagnosis System. In: International Conference on Robotics and Mechatronics (ICRoM), First RSI/ISM 2013: IEEE. pp. 205-211.

[8] Jain AK, Farrokhnia F, editors. Unsupervised texture segmentation using Gabor filters. In: IEEE International Conference on Systems, Man and Cybernetics, 1990; 4-7 Nov 1990; Los Angeles, USA. New York, NY, USA: IEEE. pp. 14-19.

[9] Hyvarinen A, Karhunen J, Oja E. Independent Component Analysis. 1st ed. New York, NY, USA: John Wiley & Sons, 2004.

[10] Lee T-W, Lewicki MS, Sejnowski TJ. ICA mixture models for unsupervised classification of non-Gaussian classes and automatic context switching in blind signal separation. IEEE Trans Pattern Anal Mach Intell 2000; 22(10): 1078-1089.

[11] Lee T-W, Girolami M, Bell AJ, Sejnowski TJ. A unifying information-theoretic framework for independent component analysis. Comput Math Appl 2000; 39(11): 1-21.

[12] Lee T-W, Girolami M, Sejnowski TJ. Independent component analysis using an extended infomax algorithm for mixed subgaussian and supergaussian sources. Neural Comput 1999; 11(2): 417-441.

[13] Roberts SJ, Penny WD, editors. Mixtures of independent component analysers. In: 21st International Conference on Artificial Neural Networks; 14-17 June 2011; Espoo, Finland. Berlin, Germany: Springer. pp. 527-534.

[14] Plumbley MD. Geometrical methods for non-negative ICA: Manifolds, Lie groups and toral subalgebras. Neurocomputing 2005; 67: 161-197.

[15] Absil P-A, Gallivan KA, editors. Joint diagonalization on the oblique manifold for independent component analysis. In: Acoustics, Speech and Signal Processing, 2006 ICASSP 2006 Proceedings 2006 IEEE International Conference on; 2006: IEEE. pp. V-V.

[16] Tek FB, Dempster AG, Kale I. Parasite detection and identification for automated thin blood film malaria diagnosis. Comput Vision Image Understanding 2010; 114(1): 21-32.

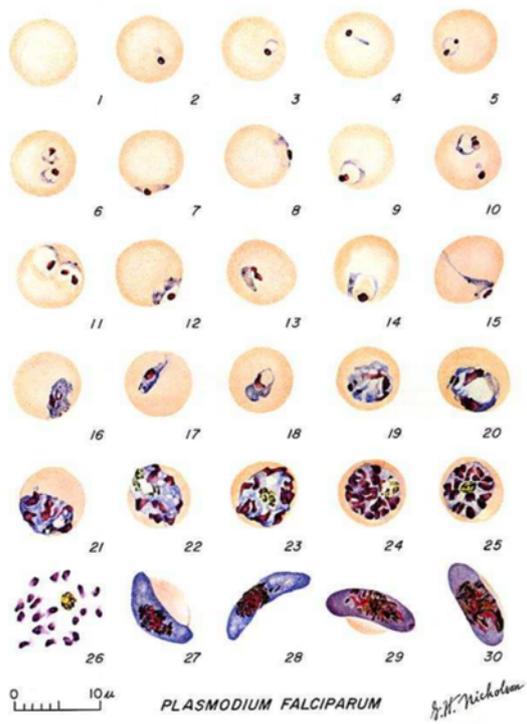

(a) 1: Normal RBC, 2-10: Rings, 11-18: Trophozoites, 19-26: Schizonts, 27-30: Gametocytes

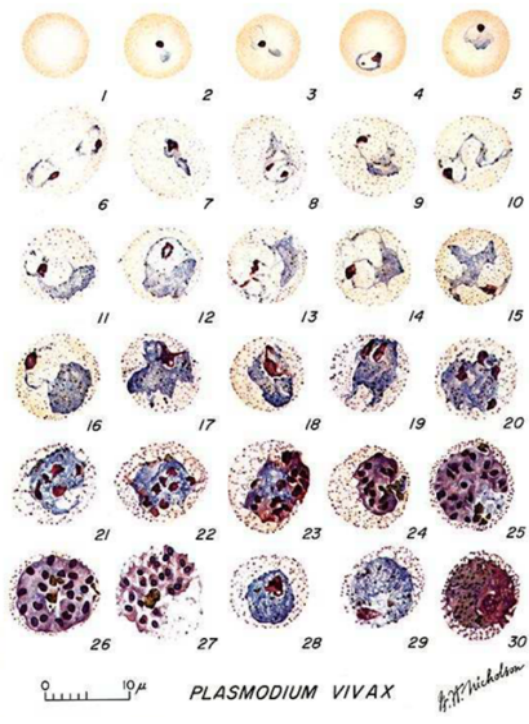

(b) 1: Normal RBC, 2-6: Rings, 7-18: Trophozoites, 19-27: Schizonts, 28-30: Gametocytes

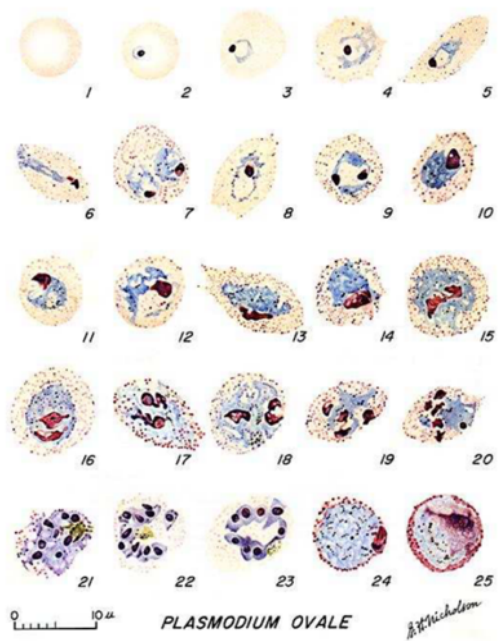

(c) 1: Normal RBC, 2-5: Rings, 6-15: Trophozoites, 16-23: Schizonts, 24-25: Gametocytes

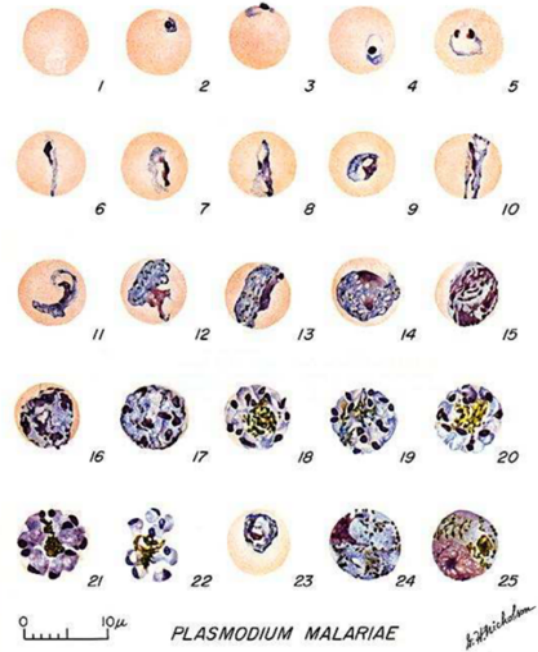

(d) 1: Normal RBC, 2-5: Rings, 6-13: Trophozoites, 14-22: Schizonts, 23-25: Gametocytes

Figure 1. Different life-stages of four species of malaria parasites (a) P. Falciparum, (b) P.Vivax, (c) P. Ovale, (d) P. Malariae

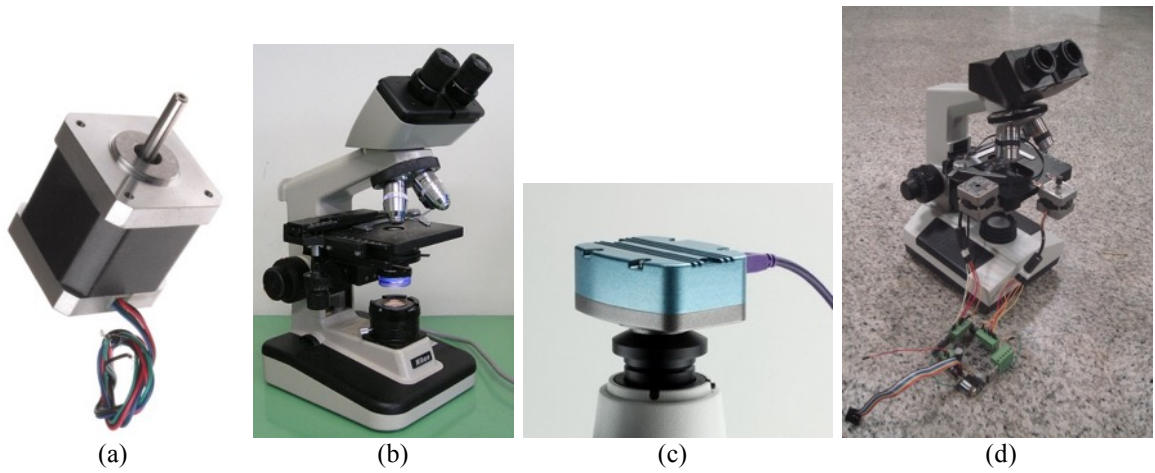

Figure 2. (a,b,c) Pieces of equipment for the automated diagnosis system (d) Complete system

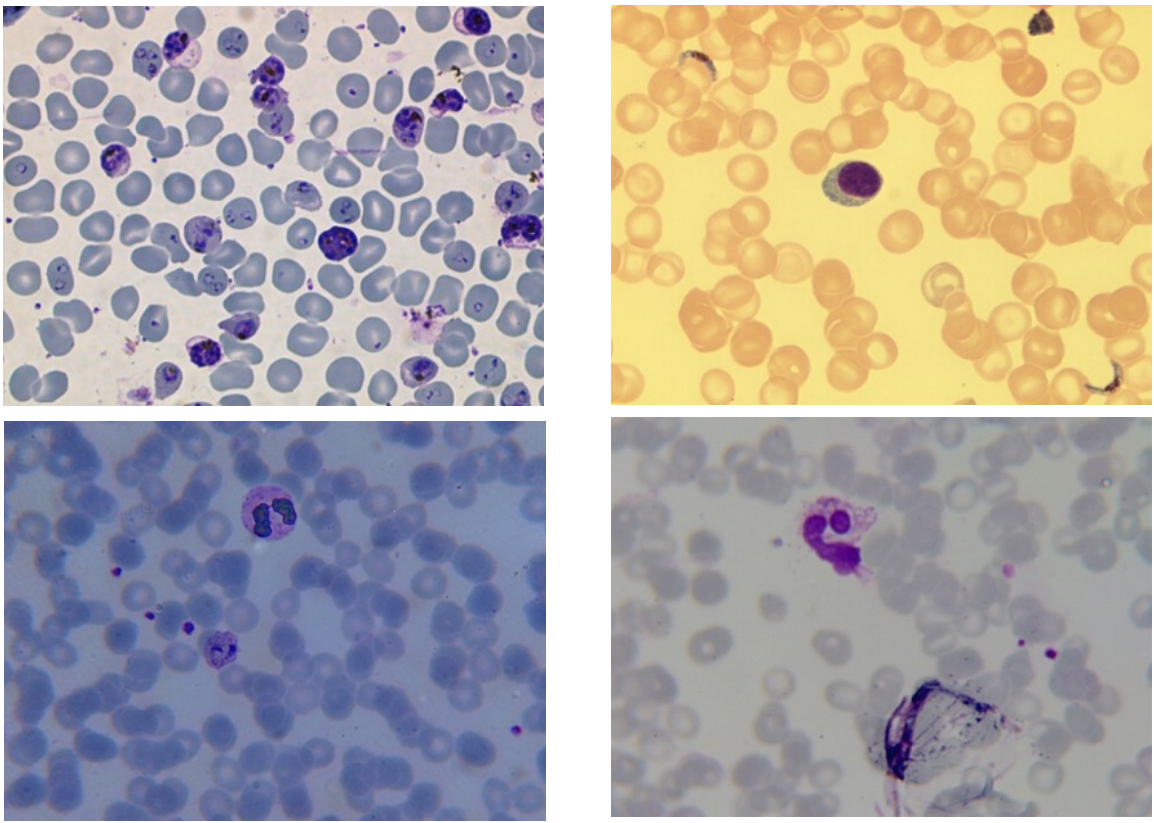

Figure 3. Different illuminations of malaria thin blood smears obtained by different microscopes

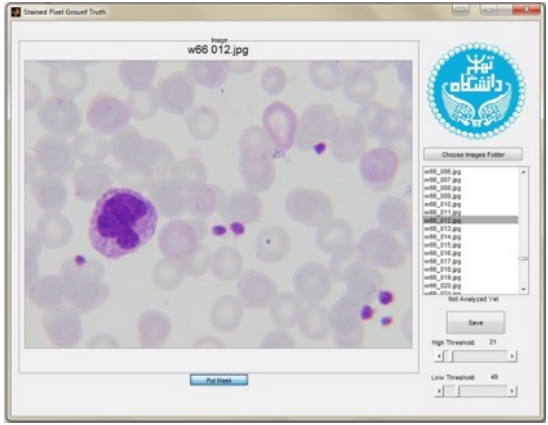
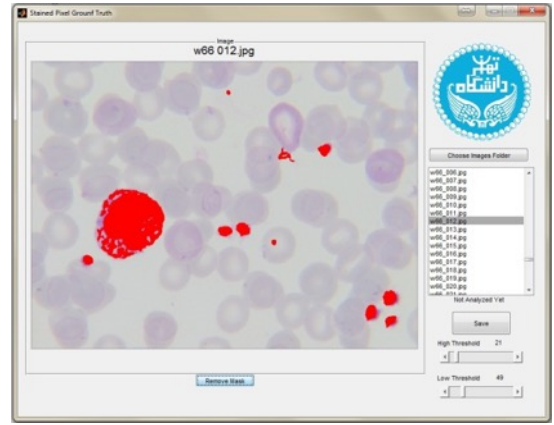

(a) A sample slide of Giemsa-stained blood smear  (b) Detected stained pixels with manual Thresholding

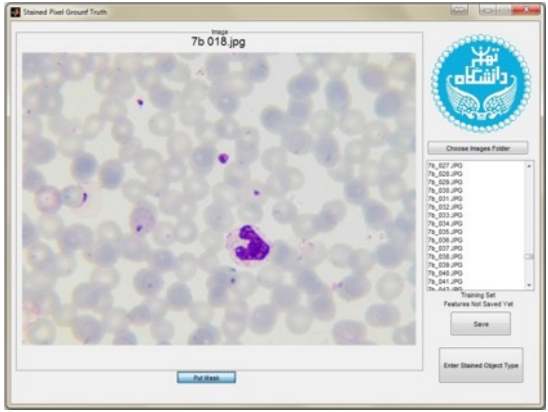
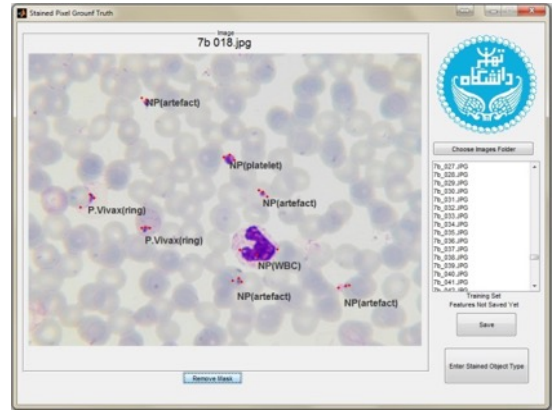

(c) A sample slide of Giemsa-stained blood smears  (d) Annotated stained objects by three-point marking

Figure 4. Two developed pripheral softwares: (a,b) To obtain ground-truth data for stained pixels (c,d) To obtain ground-truth data for staiend objects identification

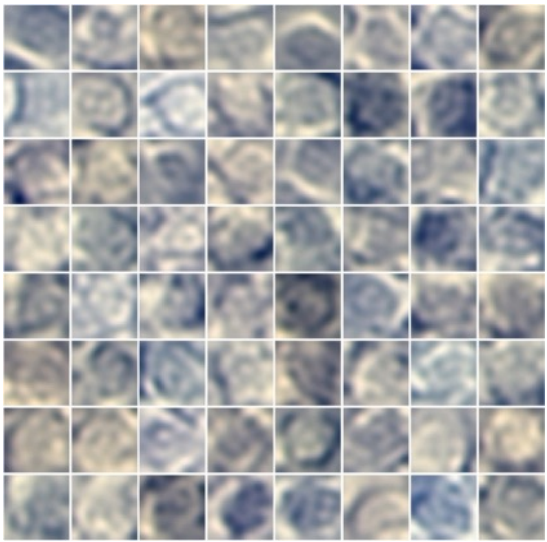
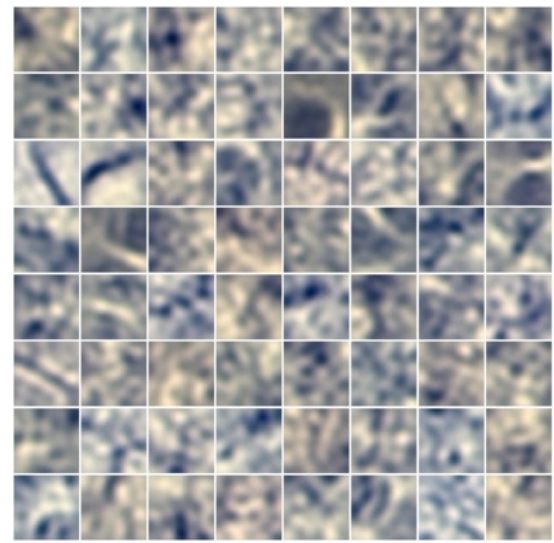

Figure 5. Features learned from the 2-component MoICA (a) Features learned as background cells (b) Features learned as foreground stained objects

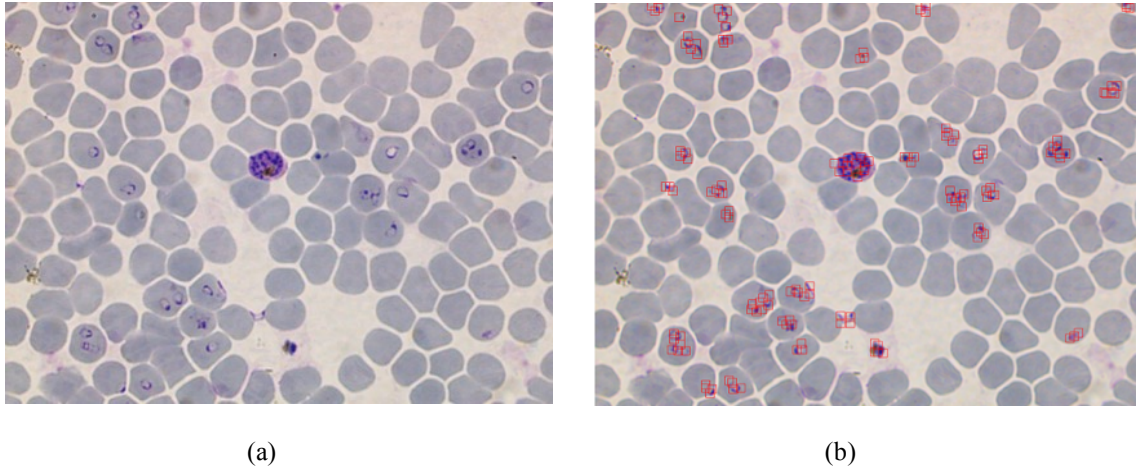

(a)                          (b)

Figure 6. (a) Blood sample image which is heavily contaminated with malaria parasites (b) the same image as (a) on which patches corresponding to one MoICA component are superimposed (the scale of the shown rectangles is not the same as the learned patches for better visibility, however they are concentric with the original patches.)

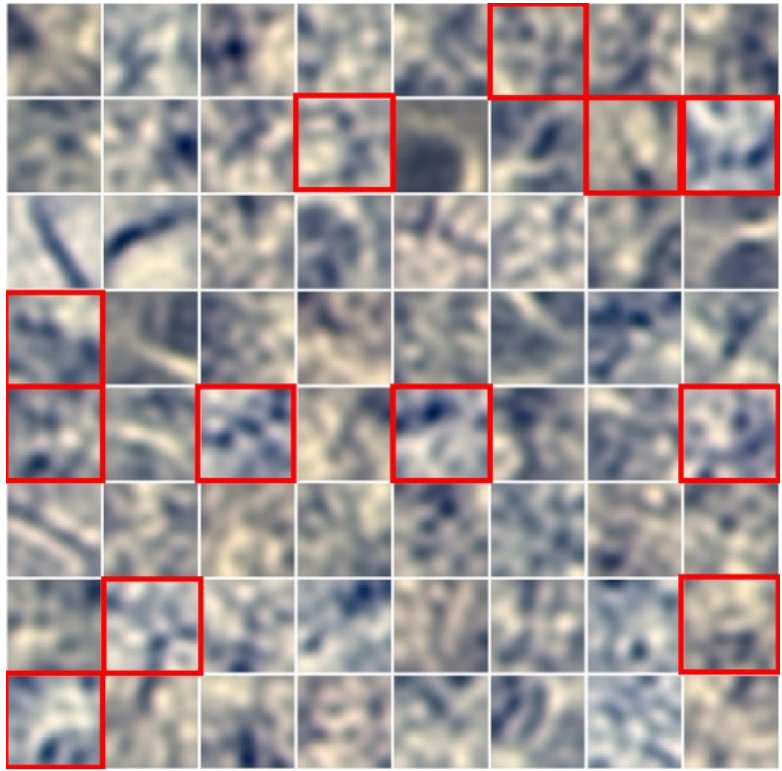

Figure 7. The same learned features as Figure 5b, though the features related to ring-shaped parasites are marked by a non-expert microscopist

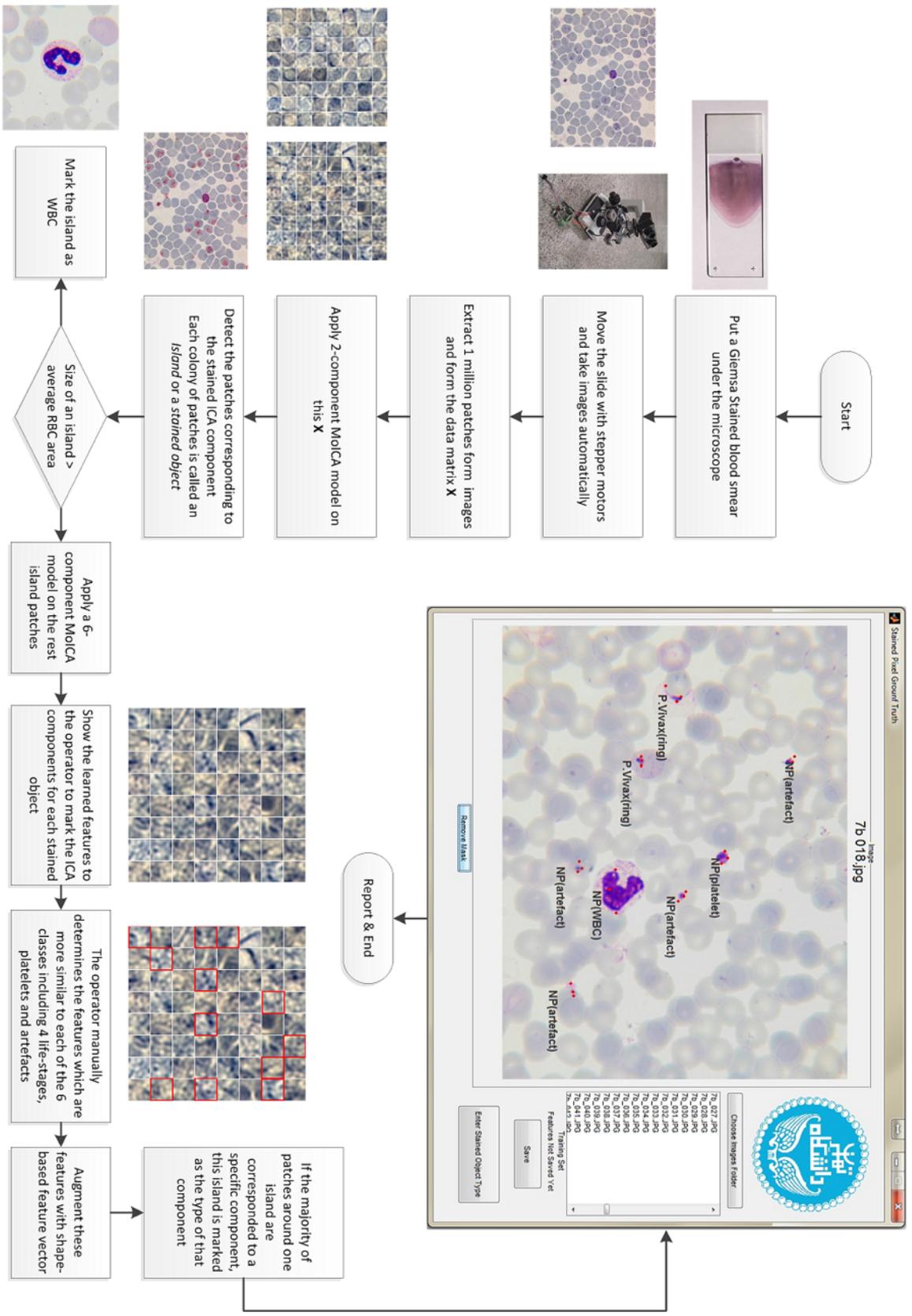

Figure 8. Overall scheme of the proposed diagnosis process

| Device | Description | Price |
|---|---|---|
| Microscope | With 100x magnification | 200 USD |
| Stepper Motors | Two instances with 0.9° accuracy | 2×20 USD |
| Control Circuit | ATMega8 microcontroller, L298 motor driver, Serial Port RS232 converter IC, PCB board | 50 USD |
| CCD Camera | Medium resolution | 200 USD |
| Overall | | less than 500 USD |

Table 1. The price and description of devices used in the diagnosis system

| Method | True Detection Rate | False Detection Rate |
|---|---|---|
| Bayes (Several supervised stages+ heuristic choice of threshold parameter) | 88.4% | 0.7% |
| MoICA (No human intervention) | 82.2% | 2.3% |

Table 2. Comparison of supervised and unsupervised methods in stained pixel classification

(a)

| | R | T | S | G | P | A |
|---|---|---|---|---|---|---|
| R | 386 | 11 | 0 | 0 | 7 | 35 |
| T | 7 | 83 | 1 | 2 | 5 | 14 |
| S | 0 | 0 | 21 | 0 | 0 | 1 |
| G | 0 | 0 | 1 | 23 | 0 | 3 |
| P | 19 | 5 | 0 | 2 | 1421 | 117 |
| A | 31 | 4 | 2 | 1 | 60 | 1418 |

(b)

| | R | T | S | G | P | A |
|---|---|---|---|---|---|---|
| R | 318 | 23 | 0 | 0 | 7 | 42 |
| T | 27 | 64 | 0 | 2 | 10 | 16 |
| S | 0 | 1 | 22 | 1 | 2 | 5 |
| G | 0 | 0 | 1 | 20 | 0 | 9 |
| P | 25 | 7 | 0 | 2 | 1377 | 135 |
| A | 73 | 8 | 2 | 3 | 87 | 1381 |

Table 3. 6-class confusion matrix for (a) The proposed unsupervised method (b) Tek's supervised method